\documentclass[10pt,twocolumn,letterpaper]{article}

\usepackage[pagenumbers]{cvpr} 

\usepackage{graphicx}
\usepackage{amsmath}
\usepackage{amssymb}
\usepackage{booktabs}

\usepackage{bbding}
\usepackage{color} 
\usepackage{multirow}

\usepackage[pagebackref,breaklinks,colorlinks]{hyperref}

\usepackage[capitalize]{cleveref}
\crefname{section}{Sec.}{Secs.}
\Crefname{section}{Section}{Sections}
\Crefname{table}{Table}{Tables}
\crefname{table}{Tab.}{Tabs.}

\def\cvprPaperID{} 
\def\confName{CVPR}
\def\confYear{2023}

\begin{document}

\title{A Light Weight Model for Active Speaker Detection}


\author{%
    Junhua Liao\textsuperscript{1}, ~~%
    Haihan Duan\textsuperscript{2}, ~~%
    Kanghui Feng\textsuperscript{1}, ~~%
    Wanbing Zhao\textsuperscript{1}, ~~%
    Yanbing Yang\textsuperscript{1,3}, ~~%
    Liangyin Chen\textsuperscript{1,3}\thanks{Corresponding author} ~~%
    \\
    {\textsuperscript{1} College of Computer Science, Sichuan University, Chengdu, China.}\\
    {\textsuperscript{2} The Chinese University of Hong Kong, Shenzhen, China.}\\
    {\textsuperscript{3} The Institute for Industrial Internet Research, Sichuan University, Chengdu, China.}\\
 {\tt\small \{liaojunhua, fengkanghui, wanbingzhao\}@stu.scu.edu.cn; }\\
 {\tt\small \{yangyanbing,  chenliangyin\}@scu.edu.cn; haihanduan@link.cuhk.edu.cn} }

\maketitle
\pagestyle{empty}  
\thispagestyle{empty} 

\begin{abstract}

Active speaker detection is a challenging task in audio-visual scenario understanding, which aims to detect who is speaking in one or more speakers scenarios. This task has received extensive attention as it is crucial in applications such as speaker diarization, speaker tracking, and automatic video editing. The existing studies try to improve performance by inputting multiple candidate information and designing complex models. Although these methods achieved outstanding performance, their high consumption of memory and computational power make them difficult to be applied in resource-limited scenarios. Therefore, we construct a lightweight active speaker detection architecture by reducing input candidates, splitting 2D and 3D convolutions for audio-visual feature extraction, and applying gated recurrent unit (GRU) with low computational complexity for cross-modal modeling. Experimental results on the AVA-ActiveSpeaker dataset show that our framework achieves competitive mAP performance (94.1\% vs. 94.2\%), while the resource costs are significantly lower than the state-of-the-art method, especially in model parameters (1.0M vs. 22.5M, about $23\times$) and FLOPs (0.6G vs. 2.6G, about $4\times$). In addition, our framework also performs well on the Columbia dataset showing good robustness. The code and model weights are available at \url{https://github.com/Junhua-Liao/Light-ASD}.

\end{abstract}

\section{Introduction}
\label{sec:intro}
Active speaker detection is a multi-modal task aiming to identify active speakers from a set of candidates in an arbitrary video. This task plays an essential role in speaker diarization~\cite{chung2020spot, wang2018speaker}, speaker tracking~\cite{qian2021audio, qian2021multi}, automatic video editing~\cite{liao2020occlusion, duan2022flad}, and other applications, which has attracted extensive attention from both industry and academia.

\begin{figure}[!t]
  \centering
  \includegraphics[width=\linewidth]{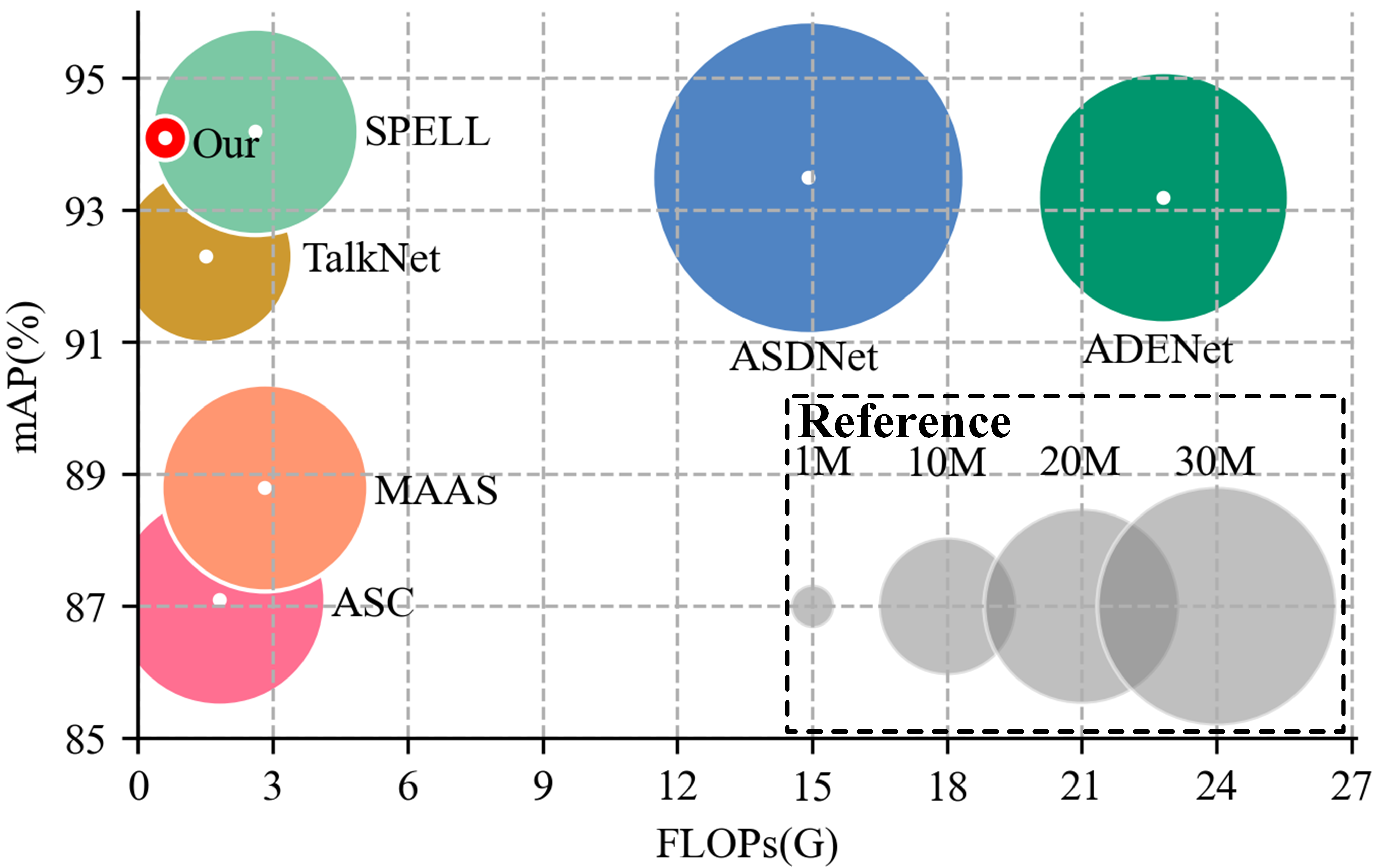}
   \caption{mAP vs. FLOPs, size $\propto$ parameters. The mAP of different active speaker detection methods~\cite{alcazar2020active, alcazar2021maas, tao2021someone, 9858007, kopuklu2021design, min2022learning} on the benchmark and the FLOPs required to predict one frame containing three candidates. The size of the blobs is proportional to the number of model parameters. The legend shows the size of blobs corresponding to the model parameters from $1\times10^6$ to $30\times10^6$.}
   \label{fig:one}
\end{figure}

The research in active speaker detection dates back more than two decades~\cite{slaney2000facesync, cutler2000look}, but the lack of large-scale reliable data has delayed the development of this field. With the release of the first large-scale active speaker detection dataset AVA-ActiveSpeaker~\cite{roth2020ava}, researchers have made a series of significant progress in this field~\cite{huang2020improved, truong2021right, tesema2022end, zhang2021unicon, tao2021someone} following the rapid development of deep learning for audio-visual tasks~\cite{michelsanti2021overview}. These studies improve the performance of active speaker detection by inputting face sequences of multiple candidates at the same time~\cite{alcazar2020active, alcazar2021maas, zhang2021unicon}, extracting visual features with 3D convolutional neural networks~\cite{kopuklu2021design, alcazar2022end, zhang2019multi}, modeling cross-modal information with complex attention modules~\cite{wuerkaixi2022rethinking, datta2022asd, 9858007}, etc, which brings higher memory and computation requirements. Therefore, existing works are difficult to be applied in scenarios requiring real-time processing with limited memory and computational resources, such as automatic video editing and live television.

In this paper, we propose a lightweight end-to-end architecture designed to detect active speakers in real time, where the improvements are conducted as the following three aspects: (a) \textbf{Single input:} inputting a single candidate face sequence with the corresponding audio; (b) \textbf{Feature extraction:} splitting the 3D convolution of visual feature extraction into 2D convolution and 1D convolution to extract spatial and temporal information respectively, and splitting the 2D convolution for audio feature extraction into two 1D convolutions to extract the frequency and temporal information respectively; (c) \textbf{Cross-modal modeling:} using gated recurrent unit (GRU)~\cite{chung2014empirical} with less calculation to replace the complex attention modules for the cross-modal modeling. According to the characteristics of this lightweight architecture, we design a novel loss function for model training. \Cref{fig:one} visualizes multiple metrics of different active speaker detection approaches. The experimental results show that our active speaker detection method (1.0M params, 0.6G FLOPs, 94.1\% mAP) significantly reduces the model size and computational cost, and its performance is still comparable to the state-of-the-art method~\cite{min2022learning} (22.5M params, 2.6G FLOPs, 94.2\% mAP) on the benchmark. Moreover, our method shows good robustness in cross-dataset testing. Finally, the single frame inference time of our method ranges from 0.1ms to 4.5ms, which is feasible for deploying in real-time applications.

The major contributions can be summarized as follows:
\begin{itemize}
\item We carry out lightweight design from three aspects of information input, feature extraction, and cross-modal modeling, and propose a lightweight and effective end-to-end active speaker detection framework. In addition, a novel loss function is designed for training.
\item Extensive experiments on AVA-ActiveSpeaker~\cite{roth2020ava}, a benchmark dataset for active speaker detection released by Google, show that our method is comparable to the state-of-the-art method~\cite{min2022learning} while still reducing model parameters by 95.6\% and FLOPs by 76.9\%.
\item In performance breakdown, our method achieves state-of-the-art performance in scenarios with different numbers of people and different face sizes.
\end{itemize}


\section{Related Work}
\label{sec:related}

The scientific community is increasingly interested in fusing multiple information sources to establish more effective joint representations~\cite{ngiam2011multimodal}. Audio-visual learning is a common multi-modal paradigm in the video field and is used to solve tasks such as audio-visual action recognition~\cite{gao2020listen, kazakos2019epic}, audio-visual event localization~\cite{hu2021class, rao2022decompose}, audio-visual synchronization~\cite{son2017lip, arandjelovic2018objects}, and audio-visual separation~\cite{jati2019neural, owens2018audio}. The active speaker detection studied in this paper is an instance of audio-visual separation.

Active speaker detection is to find out who is speaking in a video clip containing multiple speakers. This field was pioneered by Cutler and Davis~\cite{cutler2000look} in the early 2000s when they learned audio-visual correlation through time-delayed neural networks. Some subsequent studies have attempted to solve this task by capturing lips motion~\cite{saenko2005visual, everingham2009taking}. Although these studies have promoted the development of this field, the lack of large-scale benchmark data for training and testing limits the application of active speaker detection in the wild. To this end, Google introduced the first large-scale video dataset AVA-ActiveSpeaker~\cite{roth2020ava} for active speaker detection, which has injected new vitality into the field, resulting in the emergence of many novel solutions.

Alc{\'a}zar \etal~\cite{alcazar2020active,alcazar2021maas} first exploit the temporal contextual and relational contextual information from multiple speakers to handle the active speaker detection task. K{\"o}p{\"u}kl{\"u} \etal~\cite{kopuklu2021design} and Min \etal~\cite{min2022learning} follow this idea to design structures that can better model temporal and relational contexts to improve detection performance. After that, Zhang \etal~\cite{zhang2021unicon, zhang2021ictcas} additionally introduce spatial context to obtain a robust model by integrating three different contextual information. On the other hand, Tao \etal~\cite{tao2021someone} achieve superior performance by using cross-attention and self-attention modules to aggregate audio and visual features. Then, based on this work~\cite{tao2021someone}, Wuerkaixi \etal~\cite{wuerkaixi2022rethinking} and Datta \etal~\cite{datta2022asd} improve the performance by introducing positional encoding and improving the attention module. In order to better exploit the potential of the attention module, Xiong \etal~\cite{9858007} introduce a multi-modal layer normalization to alleviate the distribution misalignment of audio-visual features.

In summary, existing active speaker detection studies focus more on model performance but ignore the cost of inputting more candidates or designing more complex models. This makes their deployment scenarios require abundant resources, while the actual scenarios may not be ideal. In the field of user-generated content, TikTok and other applications provide a number of automatic video editing functions to assist users in their creation. Active speaker detection can provide more possibilities for this service. Since many users prefer to create on resource-constrained electronic devices such as mobile phones and tablets, a lightweight model is needed for deployment. In live television, active speaker detection can assist the director cut the shot to the current speaker, which requires the model to perform real-time detection. Therefore, it is worthwhile to investigate a lightweight and efficient active speaker detection framework to cope with extreme environments.


\section{Method}
\label{sec:method}

\begin{figure}[!ht]
  \centering
  \includegraphics[width=\linewidth]{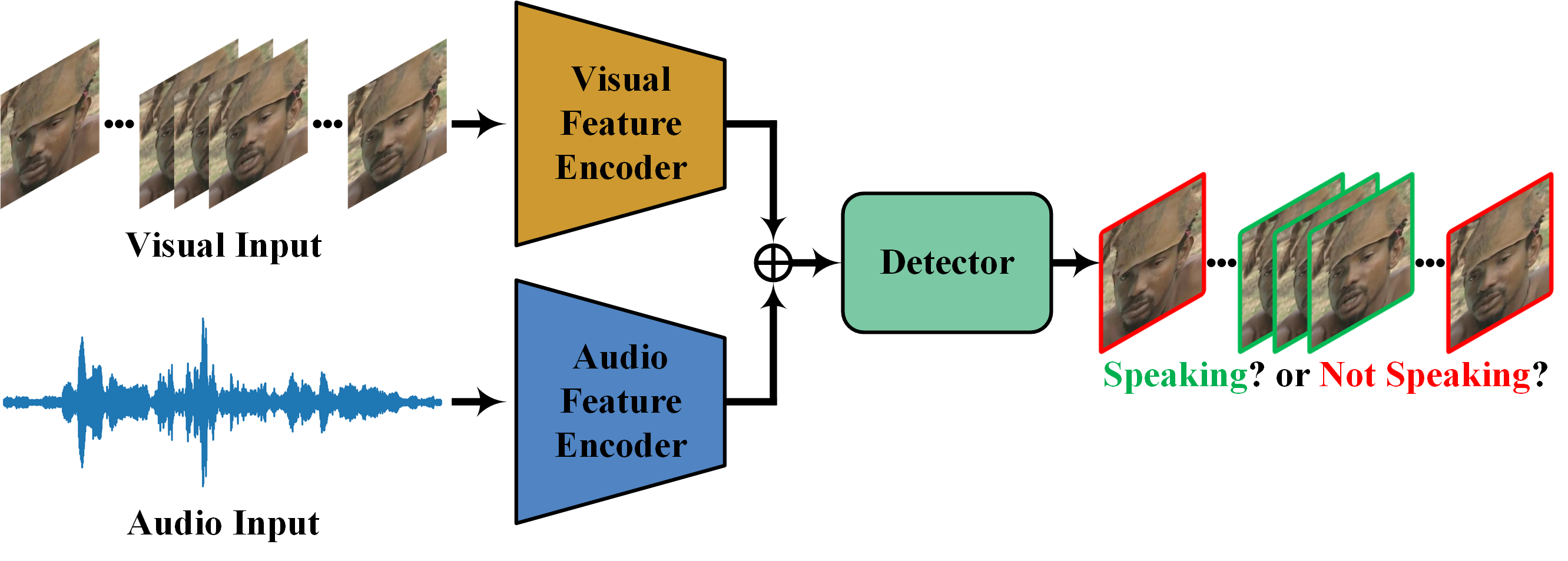}
   \caption{An overview of our active speaker detection framework.}
   \label{fig:pipeline}
\end{figure}

In this section, we describe the proposed end-to-end lightweight active speaker detection approach in detail. As shown in \cref{fig:pipeline}, our framework consists of a feature representation frontend and a speaker detection backend. The frontend contains a video feature encoder and an audio feature encoder. They encode the input candidate face sequence and the corresponding audio respectively to obtain the features of the visual signal and audio signal. In order to take full advantage of the multi-modal features, the backend detector first models the temporal context of the audio-visual features obtained by the point-wise addition of visual features and audio features, and then predicts whether the current candidate is speaking.

\subsection{Visual Feature Encoder}

\begin{figure}[!b]
  \centering
  \includegraphics[width=\linewidth]{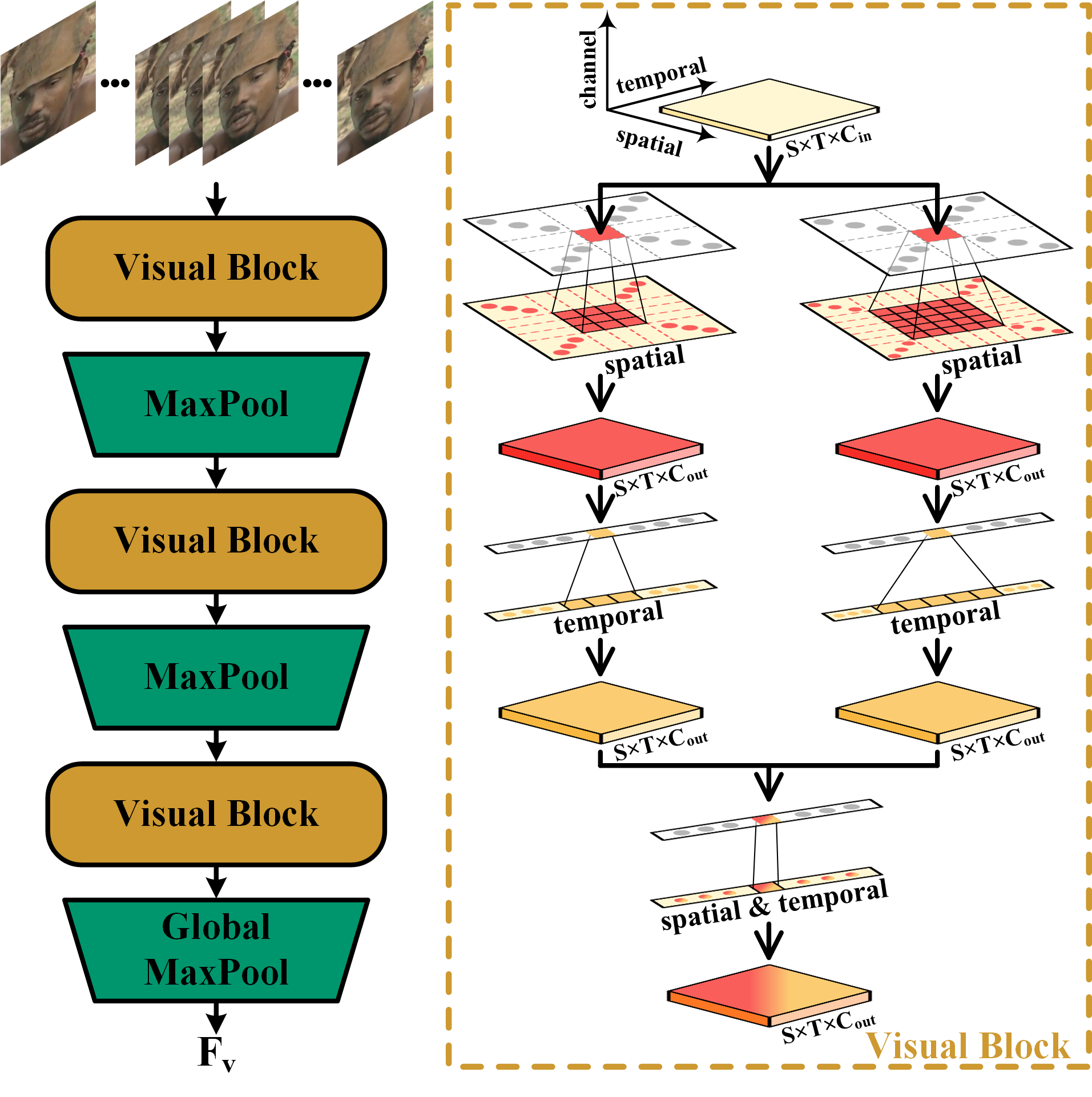}
   \caption{The architecture of visual feature encoder. The channel output dimensions $C_{out}$ of the three visual blocks are 32, 64 and 128, respectively. The MaxPool is executed in the spatial dimension, with a kernel size of 3 and stride of 2.}
   \label{fig:v_e}
\end{figure}

Many active speaker detection methods use 3D convolutional neural networks as visual feature encoders~\cite{kopuklu2021design, 9858007, zhang2019multi, alcazar2022end}. Although 3D convolution can effectively extract the spatio-temporal information of face sequences, it has a large number of model parameters and the computational cost is very expensive. In order to construct a lightweight visual feature encoder, we split the 3D convolution into 2D convolution and 1D convolution to extract the spatial and temporal information of the candidate face sequence respectively. Compared with 3D convolution, this method can greatly reduce the model parameters and computational burden while maintaining the performance~\cite{qiu2017learning, tran2018closer, liao2022light}.

Our lightweight visual feature encoder is shown in \cref{fig:v_e}. This encoder contains three visual blocks. In each visual block, there are two paths for spatio-temporal feature extraction, where one is the convolution combination after 3D convolution splitting with a kernel size of 3 and the other is the convolution combination after 3D convolution splitting with a kernel size of 5. The multiple paths are designed to extract features with different receptive fields to obtain more abundant spatio-temporal information. Then, convolution with a kernel size of 1 integrates features from different paths. Batch normalization and ReLU activation will be performed for each convolution in the visual block. It is worth noting that all convolution in the visual feature encoder has a stride of 1, except the 2D convolution in the first visual block, which has a stride of 2. This means that they reduce the spatial dimension when extracting features, which results in smaller feature maps generated by the visual feature encoder in subsequent feature extraction. The small-size feature map can not only reduce the memory footprint, but also improve the computation speed~\cite{radosavovic2020designing}. Finally, we use global max pooling in the spatial dimension to obtain the visual feature $F_v$ of the candidate face sequence.

\subsection{Audio Feature Encoder}

\begin{figure}[!t]
  \centering
  \includegraphics[width=\linewidth]{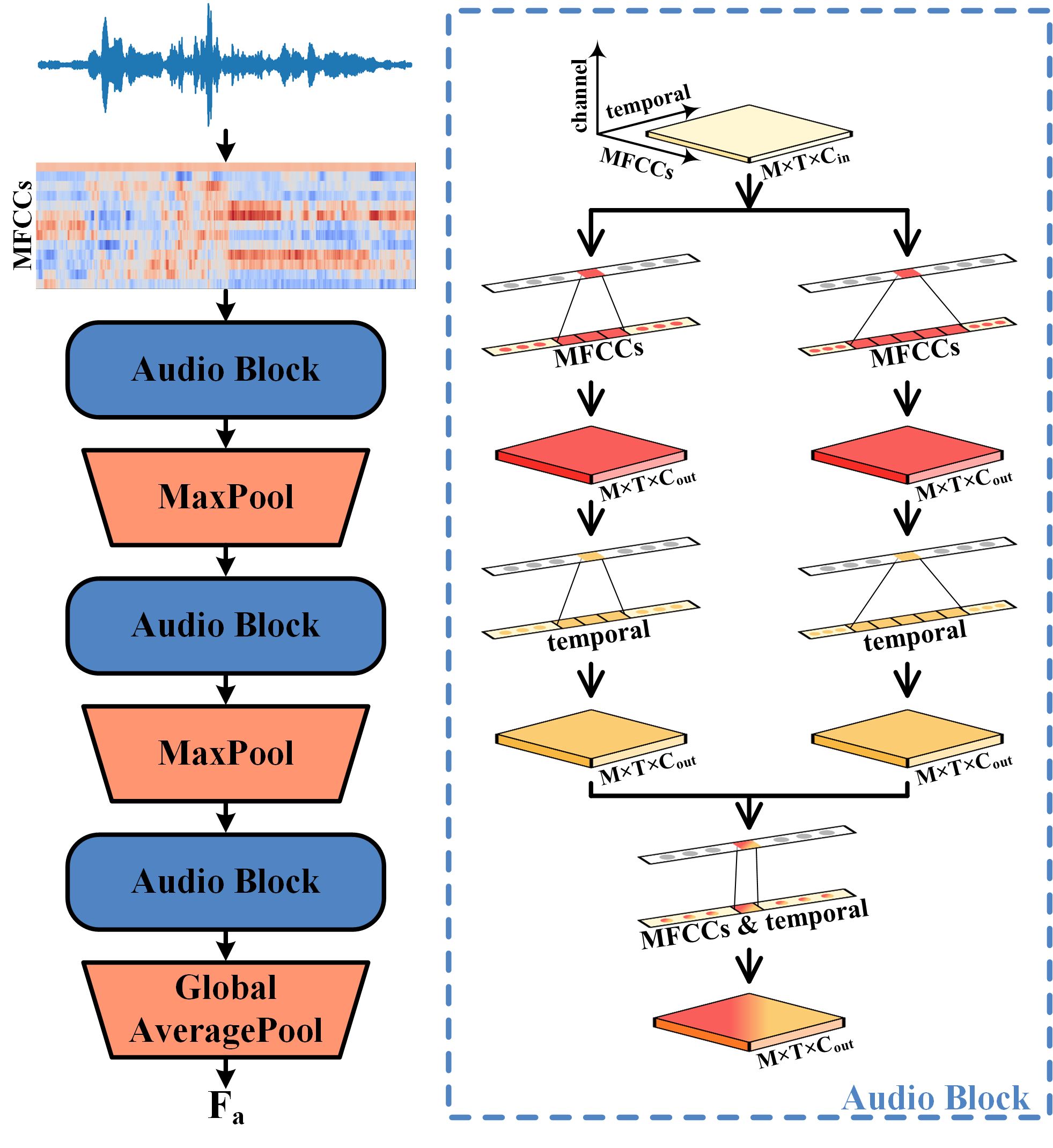}
   \caption{The architecture of the audio feature encoder. The channel output dimensions $C_{out}$ of the three audio blocks are 32, 64 and 128, respectively. The MaxPool is executed in the temporal dimension, with a kernel size of 3 and stride of 2. }
   \label{fig:a_e}
\end{figure}

Mel-frequency cepstral coefficients (MFCCs) is one of the most widely used methods in audio recognition, aiming to improve the accuracy of speech activity detection~\cite{purwins2019deep}. Therefore, like most existing active speaker detection methods~\cite{tesema2022end, wuerkaixi2022rethinking, zhang2021unicon, tao2021someone, datta2022asd, 9858007}, we extract a 2-dimensional feature map composed of 13-dimensional MFCCs and temporal information from the original audio signal as the input of the audio feature encoder. However, we do not follow the general idea of previous studies using 2D convolutional neural networks to extract audio features. In contrast, we follow the idea of lightweight in visual blocks, and split 2D convolution into two 1D convolutions to extract information from the MFCCs dimension and temporal dimension respectively. \Cref{fig:a_e} illustrates our audio feature encoder architecture, which consists of three audio blocks. Similar to the visual block, the audio block also has two paths with different receptive fields for feature extraction, and convolution with a kernel size of 1 is used for feature integration. It is worth mentioning that the first two max pooling layers in the audio feature encoder perform dimensionality reduction in the temporal dimension. Since the original audio signal sampled by the analysis window for MFCCs usually has overlapping areas between adjacent frames, pooling is required to keep the temporal dimension of audio features consistent with that of the visual features. Finally, global average pooling is performed in the MFCCs dimension to obtain the candidate's audio features $F_a$.

\subsection{Detector}
\begin{figure}[!ht]
  \centering
  \includegraphics[width=\linewidth]{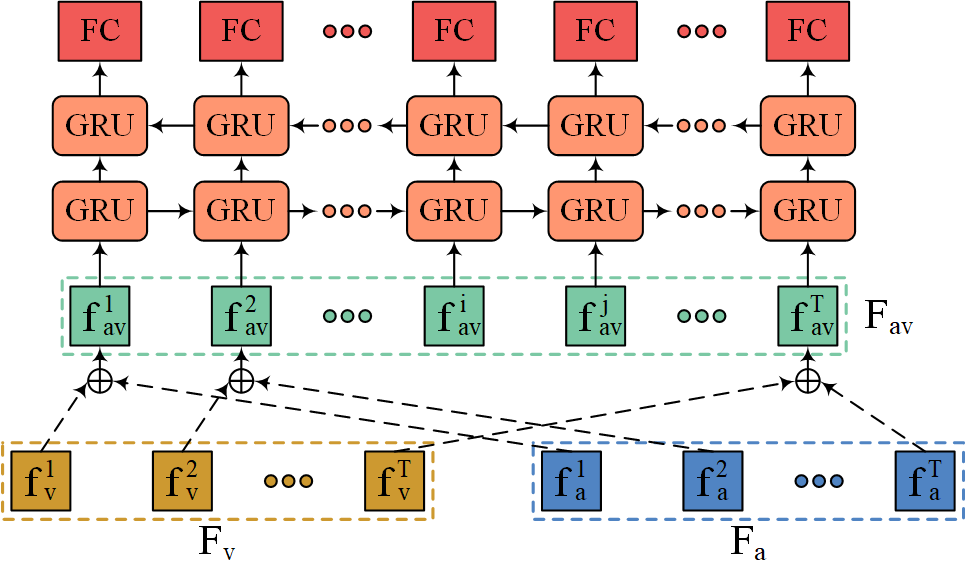}
   \caption{The architecture of the detector. $f_v^i$, $f_a^i$ and $f_{av}^i$ respectively represent the visual features, audio features, and multi-modal features of the $i_{th}$ frame of the candidate sequence.}
   \label{fig:c}
\end{figure}

We input the multi-modal features $F_{av}$ obtained by summing the visual features $F_v$ with the audio features $F_a$ into the speaker detector. The architecture of the detector is shown in \cref{fig:c}, which is also a lightweight structure. Firstly, bidirectional GRU models the temporal context information of multi-modal feature $F_{av}$. Then, a fully connected layer (FC) predicts whether the candidate speaks.

\subsection{Loss Function}

The existing active speaker detection loss function architecture usually consists of three parts: main classifier, visual auxiliary classifier, and audio auxiliary classifier~\cite{roth2020ava}. Unlike previous studies~\cite{wuerkaixi2022rethinking, zhang2021unicon, tao2021someone}, our single candidate input framework integrates visual and audio features directly without any additional cross-modal interaction. This means that the auxiliary classifiers only rely on single-modal features for prediction. In special scenarios with multiple candidates, the visual auxiliary classifier can determine whether a candidate is speaking based only on the candidate's facial information. However, without introducing visual features, the audio auxiliary classifier can only judge whether someone is speaking, but cannot judge whether the current candidate is speaking, resulting in high losses. To this end, our loss function architecture is only composed of the main classifier and the visual auxiliary classifier. 

The loss function is calculated as follows:

Firstly, we divide the prediction result by the temperature coefficient $\tau$ and perform softmax.
\begin{equation}
  p_s = \frac{exp(r_{speaking}/\tau)}{exp(r_{speaking}/\tau) + exp(r_{no\_speaking}/\tau)}
  \label{eq:one}
\end{equation}
where $r_{speaking}$ and $r_{no\_speaking}$ respectively represent the prediction result of whether the current candidate speaks, and $p_s$ denotes the probability of the candidate speaking. 

Note in particular that the temperature coefficient $\tau$ will gradually decrease during the training. 
\begin{equation}
  \tau = \tau_0 - \alpha e
  \label{eq:four}
\end{equation}
where $\tau_0$ is set to 1.3 as the initial temperature and $\alpha$ is set to 0.02 as the decay degree. $e$ indicates the training epoch.

Secondly, we calculate the loss $\mathcal{L}$.
\begin{equation}
  \mathcal{L} = -\frac{1}{T} \sum\limits_{i = 1}^T (\mathit{g}^i\log(p^i_s)+(1-g^i)\log(1-p^i_s))
  \label{eq:two}
\end{equation}
where $p^i_s$ and $g^i$ are the probability and ground truth of the candidate speaking in the $i_{th}$ frame of the video. T refers to the number of video frames.

Finally, we obtain the complete loss function $L_{asd}$.
\begin{equation}
  L_{asd} = \mathcal{L}_{av} + \lambda \mathcal{L}_{v}
  \label{eq:three}
\end{equation}
where $\mathcal{L}_{av}$ and $\mathcal{L}_{v}$ denote the loss of the main classifier and the visual auxiliary classifier respectively, and $\lambda$ is the weight coefficient set to 0.5.

\begin{table*}
  \centering
  \begin{tabular}{@{}lcccccc@{}}
    \toprule
    Method & Single candidate? & Pre-training? & E2E? & Params(M) & FLOPs(G) & mAP(\%)\\
    \midrule
    ASC (CVPR'20)~\cite{alcazar2020active} & \XSolidBrush & \Checkmark & \XSolidBrush & 23.5 & 1.8 & 87.1\\
    MAAS (ICCV'21)~\cite{alcazar2021maas} & \XSolidBrush & \Checkmark & \XSolidBrush & 22.5 & 2.8 & 88.8\\
    Sync-TalkNet (MLSP'22)~\cite{wuerkaixi2022rethinking} & \Checkmark & \XSolidBrush & \Checkmark & \textgreater14.6 & \textgreater1.5(0.5$*$3) & 89.8\\
    UniCon (MM'21)~\cite{zhang2021unicon} & \XSolidBrush & \Checkmark & \XSolidBrush & \textgreater22.4 & \textgreater1.8 & 92.2\\
    TalkNet (MM'21)~\cite{tao2021someone} & \Checkmark & \XSolidBrush & \Checkmark & 15.7 & 1.5(0.5$*$3) & 92.3\\
    ASD-Transformer (ICASSP'22)~\cite{datta2022asd} & \Checkmark & \XSolidBrush & \Checkmark & \textgreater14.6 & \textgreater1.5(0.5$*$3) & 93.0\\
    ADENet (TMM'22)~\cite{9858007} & \Checkmark & \XSolidBrush & \Checkmark & 33.2 & 22.8(7.6$*$3) & 93.2\\
    ASDNet (ICCV'21)~\cite{kopuklu2021design} & \XSolidBrush & \Checkmark & \XSolidBrush & 51.3 & 14.9 & 93.5\\
    EASEE-50 (ECCV'22)~\cite{alcazar2022end} & \XSolidBrush & \Checkmark & \Checkmark & \textgreater74.7 & \textgreater65.5 & 94.1\\
    SPELL (ECCV'22)~\cite{min2022learning} & \XSolidBrush & \Checkmark & \XSolidBrush & 22.5 & 2.6 & \textbf{94.2}\\
    \textbf{Our Method} & \Checkmark & \XSolidBrush & \Checkmark & \textbf{1.0} & \textbf{0.6}(0.2$*$3) & 94.1\\
    \bottomrule
  \end{tabular}
  \caption{Performance comparisons on the validation set of the AVA-ActiveSpeaker dataset~\cite{roth2020ava}. For each method, we copy the results from its original paper or calculate from the open-source code. Some studies~\cite{zhang2021unicon,datta2022asd,alcazar2022end,wuerkaixi2022rethinking} are not yet open source, so we only estimate the parameters and FLOPs of their audio-visual encoder. The E2E indicates end-to-end. The FLOPs indicates the number of floating point operations required to calculate one frame containing three candidates. The FLOPs of the single candidate input method will be tripled.}
  \label{tab:result}
\end{table*}

\section{Experiment}
\label{sec:exper}

\subsection{Dataset}
\textbf{AVA-ActiveSpeaker.} The AVA-ActiveSpeaker dataset \cite{roth2020ava} is the first large-scale standard benchmark for active speaker detection. It consists of 262 Hollywood movies, 120 of which are training sets, 33 are validation sets, and the remaining 109 are test sets. The entire dataset contains normalized bounding boxes for 5.3 million faces, and each face detection is assigned a speaking or non-speaking label. As a mainstream benchmark for active speaker detection tasks, this dataset contains occlusions, low resolution faces, low quality audio, and varied lighting conditions, making it highly challenging. It is worth noting that the test set is provided for the ActivityNet challenge and is not available. Therefore, we perform performance evaluation on the validation set as in previous works~\cite{tesema2022end, 9858007, datta2022asd, zhang2021unicon, wuerkaixi2022rethinking}.

\textbf{Columbia.} The Columbia dataset~\cite{chakravarty2016cross} is another standard test benchmark for active speaker detection. This dataset consists of an 87-minute panel discussion video. In the video, 5 speakers (Bell, Boll, Lieb, Long, Sick) take turns speaking, and 2-3 speakers are visible at any given time.

\subsection{Implementation Details}
We reshape every face to 112 × 112. Our final architecture is implemented by PyTorch~\cite{paszke2019pytorch} and all experiments are performed using an NVIDIA RTX 3090 GPU (24GB). These models utilize the Adam optimizer over 30 training epochs, where the learning rate is set as 0.001 with a decay rate of 0.05 every epoch. 

\textbf{Evaluation metric.} According to the common protocol, the metric of the AVA-ActiveSpeaker dataset is the mean Average Precision (mAP), and the metric of the Columbia dataset is the F1 score. Additionally, we also report model parameters and floating point operations (FLOPs) to measure different models' size and complexity.

\subsection{Comparison with the state-of-the-art}
We compare the performance of our framework with other active speaker detection methods~\cite{alcazar2020active,alcazar2021maas,zhang2021unicon,tao2021someone,datta2022asd,9858007,kopuklu2021design,alcazar2022end,min2022learning,wuerkaixi2022rethinking} on the AVA-ActiveSpeaker validation set, and summarize these results in \cref{tab:result}. We highlight four aspects of the experimental results. (a) \textbf{Lightweight and efficient}. The mAP of our framework reaches 94.1\%, slightly inferior to the 94.2\% of the state-of-the-art method SPELL\cite{min2022learning}, with 23 times fewer model parameters and 4 times less computation. (b) \textbf{End-to-End}. Our method and EASEE-50~\cite{alcazar2022end} are the state-of-the-art end-to-end active speaker detection method, with more than 75 times fewer model parameters and 66 times less computation. (c) \textbf{No pre-training}. Unlike approaches~\cite{alcazar2020active, alcazar2021maas, kopuklu2021design,zhang2021unicon,alcazar2022end,min2022learning} that use other large-scale datasets for pre-training models, our architecture only uses the AVA-ActiveSpeaker training set to train the entire network from scratch without any additional processing. (d) \textbf{Single candidate}. Existing studies~\cite{alcazar2020active, alcazar2021maas, kopuklu2021design,zhang2021unicon,alcazar2022end,min2022learning} prefer to exploit the relational contextual information between speakers to improve performance. In order to reduce the computational burden, our model only inputs a single candidate, which means that our architecture can make accurate predictions based on the visual and audio signals of a single candidate. In general, the experimental results on the benchmark dataset of active speaker detection support the effectiveness and superiority of our lightweight framework. 

Contrary to our lightweight model design philosophy, the state-of-the-art end-to-end active speaker detection method EASEE-50~\cite{alcazar2022end} uses a 3D convolutional neural network to extract the visual features of multiple input candidate face sequences. By increasing the amount of information and the complexity of the model, its performance improves to 94.1\%, but the number of model parameters and FLOPs also reaches more than 74.7M and 65.5G, respectively. Multi-candidate input amplifies the disadvantage of expensive computation of visual feature encoder based on 3D convolution, because each inference requires more computational resources to extract the visual features of multiple candidate faces. In contrast, our model achieves the same mAP using only about 1\% of the number of model parameters and computational cost of the EASEE-50. This shows that the small model can also achieve excellent performance in the active speaker detection task.

Additionally, to evaluate the robustness of our method, we also test it on the Columbia dataset~\cite{chakravarty2016cross}. The experimental results are shown in \cref{tab:col}. Without fine-tuning, our method achieves a state-of-the-art average F1 score of 81.1\% on the Columbia dataset compared with TalkNet~\cite{tao2021someone} and LoCoNet~\cite{wang2023loconet}, showing good robustness.

\begin{table}[!ht]
  \centering
  \begin{tabular}{@{}lccccc|c@{}}
    \toprule
    \multirow{2}{*}{Method} & \multicolumn{6}{c}{Speaker}\\
    & Bell & Boll & Lieb & Long & Sick & Avg\\
    \midrule
    TalkNet~\cite{tao2021someone} & 43.6 & 66.6 & 68.7 & 43.8 & 58.1 & 56.2\\
    LoCoNet~\cite{wang2023loconet} & 54.0 & 49.1 & 80.2 & \textbf{80.4} & 76.8 & 68.1\\
    \textbf{Our Method} & \textbf{82.7} & \textbf{75.7} & \textbf{87.0} & 74.5 & \textbf{85.4} & \textbf{81.1}\\
    \bottomrule
  \end{tabular}
  \caption{Comparison of F1-Score (\%) on the Columbia dataset~\cite{chakravarty2016cross}.}
  \label{tab:col}
\end{table}

\subsection{Ablation Studies}

\textbf{Kernel size.}
\begin{table}[!b]
  \centering
  \begin{tabular}{@{}lccc@{}}
    \toprule
    Kernel size & Params(M) & FLOPs(G) & mAP(\%) \\
    \midrule
    3 & 0.50 & 0.21 & 93.0 \\
    5 & 0.77 & 0.42 & 93.4 \\
    7 & 1.12 & 0.72 & 93.4 \\
    3 and 5 & 1.02 & 0.63 & 94.1 \\
    \bottomrule
  \end{tabular}
  \caption{Impact of convolutional kernel size.}
  \label{tab:ab1}
\end{table}
We evaluate the performance of the frontend feature encoder with different convolutional kernel sizes, and the results are shown in \cref{tab:ab1}. When the encoders use convolutions with a kernel size of 3, the whole framework can achieve 93.0\% mAP with only 0.5M model parameters and 0.21G FLOPs, outperforming many active speaker detection methods~\cite{alcazar2020active,alcazar2021maas,zhang2021unicon,tao2021someone,wuerkaixi2022rethinking}. When the size of the convolutional kernel increases from 3 to 5, the amount of information input in the feature extraction process increases, and the performance of the model is improved. However, when the convolutional kernel size is increased from 5 to 7, only the number of model parameters and the computation amount increase significantly, but the performance does not improve. This shows that properly increasing the receptive field is helpful to improve the model performance. In addition, we also combine convolutions with different kernel sizes and achieve the best performance of 94.1\% by combining information under different receptive fields. This verifies the rationality and effectiveness of multipath design in the visual block and the audio block.

\textbf{Visual feature encoder.}
We experimentally verify the effectiveness of our lightweight visual feature encoder, and the results are shown in \cref{tab:ab2}. Due to the expensive computational cost of 3D convolution, many active speaker detection methods use 2D convolutional neural networks to extract the spatial features of face sequences, and then use additional modules to extract temporal features~\cite{alcazar2020active, alcazar2021maas, min2022learning, zhang2021unicon,tao2021someone}. Therefore, we use the visual encoder from TalkNet~\cite{tao2021someone} in our framework to verify whether the traditional ideas are more effective. This visual encoder consists of ResNet-18~\cite{he2016deep} and a visual temporal module. It can be seen that, after the introduction of this visual encoder, the number of parameters and FLOPs of the overall architecture have reached 13.68M and 1.53G respectively, but the performance has not been improved, only 92.8\%. Although the large-capacity model can learn more knowledge, the input of this study is small and relatively simple face images, so the small model with exquisite design is enough to complete the task of feature extraction. In addition, the dimension of features extracted by ResNet is relatively high, and researchers usually reduce the dimension and then conduct multi-modal modeling, which inevitably leads to information loss. Therefore, the features extracted by our visual encoder are only 128 dimensions, which can not only meet the design concept of lightweight, but also avoid the information loss caused by dimension reduction. We also evaluate the performance when the visual blocks in our visual feature encoder use 3D convolution. Although the encoder is lightweight, 3D convolution still doubles the number of model parameters and FLOPs without any performance improvement. Compared with 3D convolution, the combination of 2D convolution and 1D convolution doubles the number of nonlinear rectifications, allowing the model to represent more complex functions. Therefore, reasonably splitting 3D convolution is not only conducive to model lightweight, but also improves model performance.

\begin{table}[!ht]
  \centering
  \begin{tabular}{@{}lccc@{}}
    \toprule
    Encoder & Params(M) & FLOPs(G) & mAP(\%) \\
    \midrule
    TalkNet~\cite{tao2021someone} & 13.68 & 1.53 & 92.8 \\
    3D convolution & 2.06 & 1.56 & 92.9 \\
    Our Method & 1.02 & 0.63 & 94.1 \\
    \bottomrule
  \end{tabular}
  \caption{Impact of visual feature encoder.}
  \label{tab:ab2}
\end{table}

\textbf{Audio feature encoder.}
\Cref{tab:ab3} shows the performance of our active speaker detection framework using different audio feature encoders. Since the audio feature map is a 2-dimensional signal composed of MFCCs and temporal information, many active speaker detection methods use ResNet-18 to extract audio features~\cite{alcazar2020active, alcazar2021maas, zhang2021unicon, datta2022asd}. Therefore, we first verify the performance of the audio encoder based on ResNet-18. The number of parameters in our framework is up to 11.98M after adopting this encoder. However, the large-capacity model seems not to provide any performance improvement, probably for similar reasons to the poor performance of ResNet in visual encoders. The large model may be prone to overfitting when extracting information from the feature map of small dimensions. Secondly, we evaluate the performance when using 2D convolution in audio blocks. Since our audio encoder is small, the number of model parameters and FLOPs shows less difference before and after 2D convolution splitting. The results show that the performance of the audio encoder based on 2D convolution is still inferior to the audio encoder based on 1D convolution by splitting. Perhaps the audio feature map does not have a strong spatial logic similar to images, so processing the MFCCs dimension and temporal dimension separately is more conducive to audio information aggregation.

\begin{table}[!ht]
  \centering
  \begin{tabular}{@{}lccc@{}}
    \toprule
    Encoder & Params(M) & FLOPs(G) & mAP(\%) \\
    \midrule
    ResNet-18~\cite{he2016deep} & 11.98 & 0.69 & 93.4 \\
    2D convolution & 1.12 & 0.63 & 93.6 \\
    Our Method & 1.02 & 0.63 & 94.1 \\
    \bottomrule
  \end{tabular}
  \caption{Impact of audio feature encoder.}
  \label{tab:ab3}
\end{table}

\textbf{Detector.}
In \cref{tab:ab4}, we show the impact of the detector using different methods to process audio-visual features on the model performance. If we directly use the FC for prediction without any processing of audio-visual features, its mAP is only 88.0\%. When we use forward GRU for temporal modeling of audio-visual features, its mAP increases by 4.6\%. This indicates that the temporal context information of audio-visual features is helpful to improve the performance of the active speaker detection model. However, the forward GRU can only transmit temporal information in one direction, which makes the amount of information obtained in each frame of the sequence unbalanced. Therefore, we use bidirectional GRU to make each frame can combine the information of the whole sequence for prediction, and achieve the best performance of 94.1\%. In addition, we also use the transformer~\cite{vaswani2017attention} as an attention module to extract temporal context information of audio-visual features, and its performance is 1.1\% lower than forward GRU. In this attention module, all frames in the sequence have the same chance to influence the current detection frame. Although this is a good mechanism, in this task, the information of the frames near the current detection frame is more helpful in determining whether the candidate is speaking. The forgetting mechanism of GRU makes neighboring frames more informative, so GRU is the better choice in this scenario.

\begin{table}[!ht]
  \centering
  \begin{tabular}{@{}lccc@{}}
    \toprule
    Detector & Params(M) & FLOPs(G) & mAP(\%) \\
    \midrule
    None & 0.82 & 0.63 & 88.0 \\
    Transformer~\cite{vaswani2017attention} & 1.02 & 0.63 & 91.5\\
    Forward GRU & 0.92 & 0.63 & 92.6 \\
    Bidirectional GRU & 1.02 & 0.63 & 94.1 \\
    \bottomrule
  \end{tabular}
  \caption{Impact of detector.}
  \label{tab:ab4}
\end{table}

\textbf{Loss function.}
\Cref{tab:ab5} shows the experimental results of whether our loss function $L_{asd}$ is helpful for model training. First of all, when our model is trained with the standard binary cross-entropy, it reaches 93.1\% mAP on the benchmark. Its performance outperforms most existing active speaker detection methods~\cite{alcazar2020active,alcazar2021maas,tesema2022end,wuerkaixi2022rethinking,zhang2021unicon,tao2021someone,datta2022asd}, which proves the superiority of our framework. After introducing our loss function $L_{asd}$ for training, the performance of the model has improved by 1\% to 94.1\%, which indicates that the visual auxiliary classifier can help us better supervise the visual feature encoder. At the same time, the introduction of temperature coefficients allows the model to avoid falling into local optimum and have more opportunities for exploration in the early stage of training, and helps the model pay more attention to difficult samples in the later stage of training to further improve accuracy.
\begin{table}[!ht]
  \centering
  \begin{tabular}{@{}lccc@{}}
    \toprule
    Method & Params(M) & FLOPs(G) & mAP(\%) \\
    \midrule
    Our (without $L_{asd}$) & 1.02 & 0.63 & 93.1 \\
    Our (with $L_{asd}$) & 1.02 & 0.63 & 94.1 \\
    \bottomrule
  \end{tabular}
  \caption{Impact of loss function.}
  \label{tab:ab5}
\end{table}

\textbf{Detection speed.}
\begin{table}[!b]
  \centering
  \begin{tabular}{@{}lcc@{}}
    \toprule
    Video frames & Inference time(ms) & FPS \\
    \midrule
    1 (about 0.04 seconds) & 4.49 & 223 \\
    500 (about 20 seconds) & 50.28 & 9944 \\
    1000 (about 40 seconds) & 96.04 & 10412 \\
    \bottomrule
  \end{tabular}
  \caption{Impact of the number of frames on the detection speed.}
  \label{tab:ab6}
\end{table}
Our active speaker detection framework supports dynamic length video input, so we evaluate the inference time and frames per second (FPS) of the model with different numbers of input frames. The specific experimental results are shown in \cref{tab:ab6}. With the increase of video frames, the performance of GPU is fully utilized and the inference time is greatly reduced. Our framework takes 96.04ms to infer 1000 frames (about 40 seconds) of video on an NVIDIA RTX 3090 GPU, while the EASEE-50~\cite{alcazar2022end} takes 2068.95ms for the audio-visual encoder portion alone. Even in the extreme case of single frame input, the inference time of our framework does not exceed 4.5ms, and the FPS can reach 223, while the FPS of EASEE-50 is less than 36. This shows that our active speaker detection method not only meets the real-time detection requirements under different input lengths, but also has a faster detection speed than the state-of-the-art end-to-end method~\cite{alcazar2022end}.

\subsection{Qualitative Analysis}

\begin{figure}[!b]
  \centering
  \begin{subfigure}{\linewidth}
    \includegraphics[width=\linewidth]{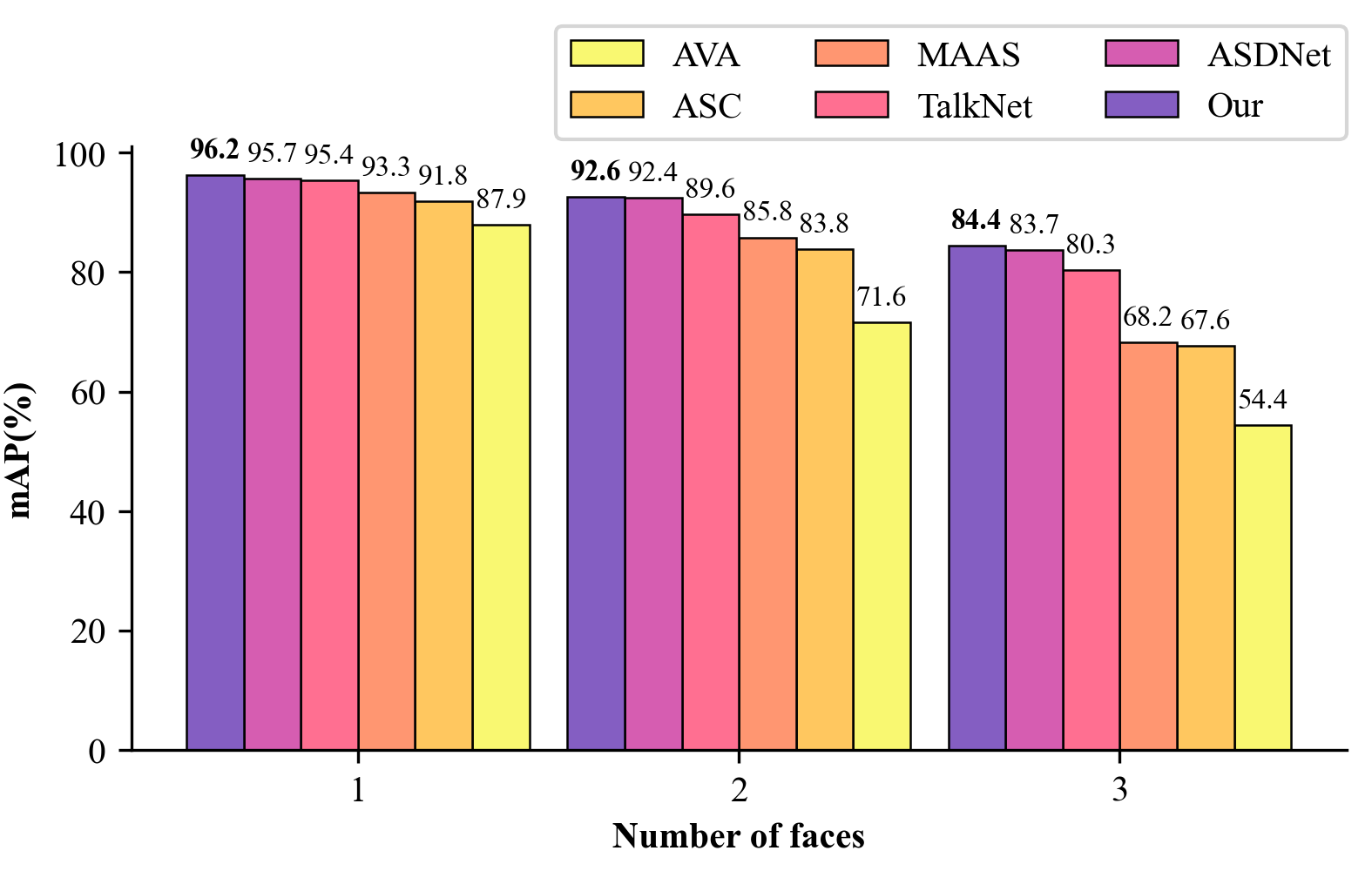}
    \caption{Performance comparison by the number of faces on each frame.}
    \label{fig:analy-a}
  \end{subfigure}
  \hfill
  \begin{subfigure}{\linewidth}
    \includegraphics[width=\linewidth]{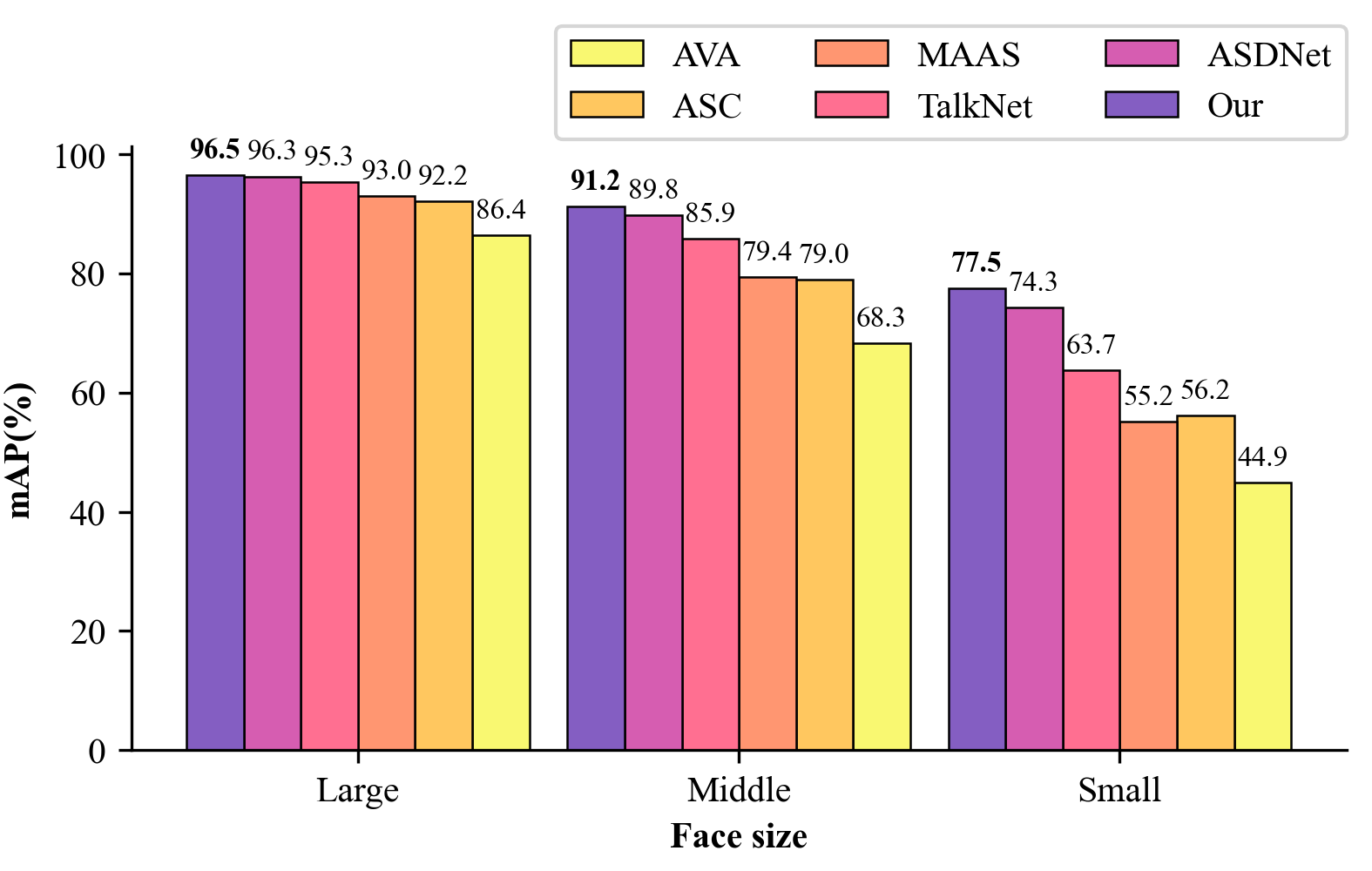}
    \caption{Performance comparison by face size.}
    \label{fig:analy-b}
  \end{subfigure}
  \caption{Performance breakdown. We evaluate the performance of our active speaker detection method and the previous state-of-the-art methods on frames with one, two, and three detected faces and on faces of different sizes.}
  \label{fig:analy}
\end{figure}

On the benchmark AVA-ActiveSpeaker, we break down the performance of our method according to the number and size of faces, just like the state-of-the-art methods~\cite{alcazar2020active, alcazar2021maas, roth2020ava, tao2021someone, kopuklu2021design}. The experimental results are shown in \cref{fig:analy}. 

First, we divide the data into three mutually exclusive groups according to the number of faces detected in the frame, which account for about 90\% of the entire validation set. \Cref{fig:analy-a} reports the performance of the active speaker detection methods based on the number of faces detected in the frame, and it can be seen that the performance of all methods decreases as the number of faces increases. Although our method only inputs one candidate per step to reduce the computational complexity, our performance consistently outperforms state-of-the-art methods with multiple inputs (ASC~\cite{alcazar2020active}, MAAS~\cite{alcazar2021maas}, and ASDNet~\cite{kopuklu2021design}) for different numbers of detected faces. The advantage of inputting multiple candidates is that the model can use not only audio-visual information but also additional relational context information to select the most likely speaker from multiple candidates. However, for the method of inputting a single candidate, the decision can only be made according to the audio-visual information of the current candidate, which has high requirements for the reliability of the audio-visual features. This confirms that our active speaker detection method can effectively extract and utilize audio-visual features to make accurate predictions.

\Cref{fig:analy-b} shows the performance of the active speaker detection methods for different face sizes. We divide the verification set into three parts according to the width of the detected faces: large (face with width more than 128 pixels), middle (face with width between 64 and 128 pixels), and small (face with width less than 64 pixels). Although the performance of all methods decreases with the decrease in face size, the advantage of our method is more significant (+0.2\% mAP, +1.4\% mAP, +3.2\% mAP). Our active speaker detection method achieves the best performance in the six scenarios subdivided by previous work, and is the only one that keeps the mAP greater than 90\% when the number of candidates is less than 3 or the face width is larger than 64 pixels, indicating that our method is significantly more robust than other competing methods.


\section{Conclusion}
\label{sec:conclu}
In this paper, we propose a lightweight framework for active speaker detection based on different aspects compared with previous works which need complex models with a large number of inputs. The key features of the proposed architecture are to input a single candidate, split 3D convolution for extracting visual features and 2D convolution for extracting audio features, and use simple modules for cross-modal modeling. Experimental results on the benchmark dataset AVA-ActiveSpeaker~\cite{roth2020ava} show that our method reduces model parameters by 95.6\% and FLOPs by 76.9\% compared to state-of-the-art methods with mAP lagging by only 0.1\%. In addition, our method shows good robustness.



\section{Acknowledgments}
\label{sec:ack}

This work is supported in part by the National Natural Science Foundation of China (Grant 62072319); in part by the Sichuan Science and Technology Program (Grant 2022YFG0041).


{\small
\bibliographystyle{ieee_fullname}
\bibliography{paper}
}

\end{document}